\documentclass[10pt]{article} 
\usepackage[accepted]{tmlr}

\usepackage{amsmath,amsfonts,bm}

\def\eqref#1{equation~\ref{#1}}

\def\1{\bm{1}}

\DeclareMathAlphabet{\mathsfit}{\encodingdefault}{\sfdefault}{m}{sl}
\SetMathAlphabet{\mathsfit}{bold}{\encodingdefault}{\sfdefault}{bx}{n}

\DeclareMathOperator*{\argmin}{arg\,min}

\usepackage{hyperref}
\usepackage{url}
\usepackage[linesnumbered,algoruled]{algorithm2e}
\usepackage{mathtools}
\usepackage{multirow}
\usepackage{amsfonts}
\usepackage{booktabs}
\usepackage{cleveref}
\usepackage{authblk}
\usepackage{xcolor}

\title{SAIF: Sparse Adversarial and Imperceptible Attack Framework}

\author[\space\space]{\textbf{Tooba Imtiaz} \thanks{Corresponding author. imtiaz.t@northeastern.edu.}}
\author[\space]{\textbf{Morgan R. Kohler}
\thanks{kohler.r.morgan@gmail.com. Work done while author was at Northeastern University.}}
\author[\space]{\textbf{Jared F. Miller}
\thanks{jared.miller@imng.uni-stuttgart.de. Work done while author was at Northeastern University.}}
\author[\space]{\textbf{Zifeng Wang}
\thanks{zifengw@google.com. Work done while author was at Northeastern University.}}
\author[\space]{\textbf{Masih Eskandar, Mario Sznaier, Octavia Camps}}
\author[]{\textbf{Jennifer Dy}
\thanks{\{eskandar.m, m.sznaier, o.camps, j.dy\}@northeastern.edu.}}

\affil[\space]{Department of Electrical \& Computer Engineering,
Northeastern University, Boston MA.}

\def\month{07}  
\def\year{2025} 
\def\openreview{\url{https://openreview.net/forum?id=YZL29eJ5j1}} 

\begin{document}

\maketitle

\begin{abstract}
 Adversarial attacks hamper the decision-making ability of neural networks by perturbing the input signal. For instance, adding calculated small distortions to images can deceive a well-trained image classification network. In this work, we propose a novel attack technique called \textbf{S}parse \textbf{A}dversarial and \textbf{I}mperceptible Attack \textbf{F}ramework (SAIF). Specifically, we design imperceptible attacks that contain low-magnitude perturbations at a few pixels and leverage these sparse attacks to reveal the vulnerability of classifiers. We use the Frank-Wolfe (conditional gradient) algorithm to simultaneously optimize the attack perturbations for bounded magnitude and sparsity with $O(1/\sqrt{T})$ convergence. Empirical results show that SAIF computes highly imperceptible and interpretable adversarial examples, and largely outperforms state-of-the-art sparse attack methods on ImageNet and CIFAR-10. Implementation of SAIF is available at \href{https://github.com/toobaimt/SAIF}{https://github.com/toobaimt/SAIF}.
\end{abstract}
\section{Introduction}
\label{sec:intro}

Deep neural networks (DNNs) are widely utilized for various tasks such as object detection \citep{redmon2016you,girshick2015fast}, classification \citep{krizhevsky2012imagenet, he2016deep}, and anomaly detection \citep{chandola2009anomaly}. These DNNs are 
ubiquitously integrated into real-world systems for medical diagnosis, autonomous driving, surveillance, etc., where misguided decision-making can have catastrophic consequences.
Therefore, it is crucial to inspect the limitations of DNNs before deployment in such safety-critical systems.

\begin{figure}[htp]
    \begin{center}
    \includegraphics[width=0.99\linewidth]
    {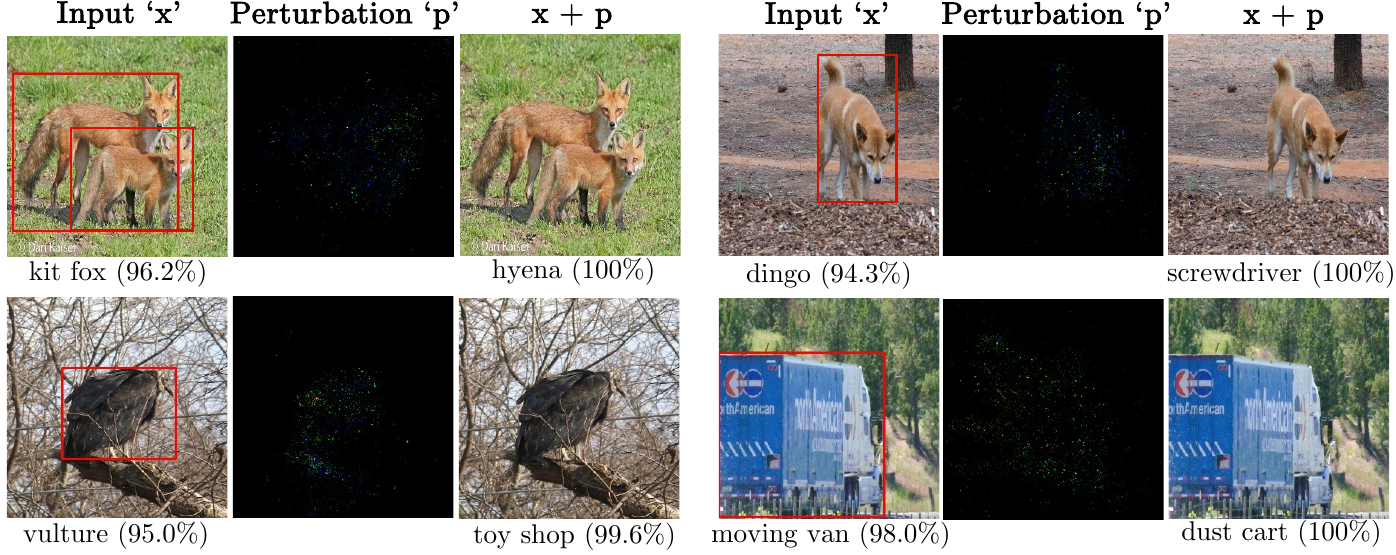}
    \end{center}
    \caption{Using the Frank-Wolfe algorithm to jointly constrain the perturbation magnitude and sparsity, we craft a highly sparse and imperceptible adversarial attack.
    By restricting attack sparsity, we can visualize the most vulnerable pixels in an image. 
    The GT bounding boxes for the subject of the input $\mathbf{x}$ are drawn in red. Note that SAIF mostly distorts pixels within that region. Inception-v3 is used for predicting labels.
    }
    \label{fig:teaser}
\end{figure}

Adversarial attacks \citep{42503} are one means of exposing the fragility of DNNs. In the classification task, these attacks can fool well-trained classifiers to make arbitrary (untargeted) \citep{moosavi2016deepfool} or targeted misclassifications \citep{carlini2017towards} by negligibly manipulating the input signal. For instance, a road sign classifier can be led to interpret a slightly modified stop sign as a speed limit sign \citep{benz2020double}.
Such adversarial attacks fool learning algorithms with high confidence while being imperceptible to the human eye. Most attack methods achieve this by constraining the pixel-wise magnitude of the perturbation. Minimizing the number of modified pixels is another strategy for making the perturbations unnoticeable \citep{su2019one, narodytska2017simple}. Bringing these two together generates high-stealth attacks \citep{modas2019sparsefool, croce2019sparse, dong2020greedyfool, fan2020sparse, williams2023black}. 

Existing methods, however, fail to produce attacks with simultaneously  \emph{very} high sparsity and low magnitude perturbations.
In this paper, we address this limitation by designing a novel method that produces strong adversarial attacks with a significantly low perturbation strength and high sparsity.
Our proposed approach, we call \textbf{S}parse \textbf{A}dversarial and \textbf{I}mperceptible attack \textbf{F}ramework (SAIF), minimally modifies only a fraction of pixels to generate highly concealed adversarial attacks. 

SAIF aims to jointly minimize the perturbation magnitude and sparsity. We formulate this objective as a constrained optimization problem. Previous works propose projection-based methods (such as PGD \citep{madry2018towards}) to optimize similar objectives, however, these require a projection step at each iteration to obtain feasible solutions \citep{croce2019sparse}. Such projections give rise to iterates very close to/at the constraint boundary, and projecting the solutions can diminish their `optimality'. 
Optimization methods such as ADMM \citep{xu2019structured, fan2020sparse} and homotopy \citep{fan2020sparse} have also been explored but have prohibitively long running times for large images.

To address these limitations, we propose to optimize our objective using the Frank-Wolfe algorithm (FW) \citep{frank1956algorithm}. FW is a projection-free, iterative method for solving constrained convex optimization problems using conditional gradients. In contrast to PGD attacks, the absence of a projection step allows Frank-Wolfe to find perturbations well within the constraint boundaries. Throughout optimization, the iterates are within the constraint limits as they are convex combinations of feasible points. Moreover, there are several algorithmic variants of Frank-Wolfe for efficient optimization.

Furthermore, the benefit from adversarial examples can be maximized by examining the vulnerabilities of deep networks alongside model explanations \citep{ignatiev2019relating, pmlr-v162-wang22e, xu2019structured}. Magnitude-constrained attacks distort all image pixels, leaving little room to interpret the additive perturbations. Our proposed attack offers explicit control over the sparsity of the distortions. This facilitates more controlled and straightforward semantic analyses, such as identifying the top-`$k$' pixels critical to fool DNNs.

Concretely, the contributions of this paper are:
\begin{itemize}
    \item We introduce a novel optimization-based adversarial attack that is visually imperceptible due to low-magnitude distortions to a fraction of image pixels.
    \item We show through comprehensive experiments that, for tight sparsity and magnitude constraints, SAIF outperforms state-of-the-art sparse attacks by a large margin (by $\geq 2\times$ higher fooling rates for most thresholds).
    \item Our sparse attack provides transparency by indicating vulnerable pixels in images. We quantitatively evaluate the overlap between perturbations and salient image regions to emphasize the utility of SAIF for such analyses.
\end{itemize}

\section{Related Works}\label{sec:related-works}

\paragraph{Magnitude-Constrained Adversarial Attacks.}
The first discovered adversarial attack by \cite{42503} uses box-constrained L-BFGS to minimize the $\ell_2$ norm of additive distortion, however, it is slow and does not scale to larger inputs. To overcome speed limitations, the Fast Gradient Sign Method (FGSM) \citep{goodfellow2014explaining} uses $\ell_{\infty}$ constrained gradient ascent w.r.t. the loss gradient for each pixel, to compute an efficient attack but with poor convergence. 
Projected Gradient Descent (PGD) is another optimization-based attack algorithm that is fast and computationally cheap, but yields solutions closer to the boundary and often fails to converge \citep{madry2018towards}. Auto-attack \citep{croce2020reliable} addresses the convergence limitations of PGD. Nevertheless, these attacks distort all the image pixels, potentially leading to high visual perceptibility. We address this shortcoming by explicitly constraining the sparsity of the adversarial perturbations.

\paragraph{Sparsity-Constrained Adversarial Attacks.}
The Jacobian-based Saliency Map Attack (JSMA) denotes pixel-wise saliency by backpropagated gradient magnitudes, then searches over the most salient pixels for a sparse targeted perturbation \citep{papernot2016limitations}. This attack is slow and fails to scale to larger images.
SparseFool \citep{modas2019sparsefool} extends DeepFool \citep{moosavi2016deepfool} to a sparse attack within the valid pixel magnitude bounds. The attack, however, is untargeted.
\citet{croce2019sparse} devise a black-box attacks PGD $\ell_0 + \ell_\infty$ and PGD $\ell_0 + \sigma$ by evaluating the impact of each pixel on logits and randomly sampling salient pixels to find a feasible sparse combination. Similar to JSMA, these attacks are expensive to compute, visually noticeable, and unstructured. The same limitations hold for SA-MOO \citep{williams2023black}.
StrAttack \citep{xu2019structured} uses ADMM (Alternating Direction Method of Multipliers) to optimize for group sparsity and perturbation magnitude constraints, but the perturbations are computationally expensive, visually noticeable, and of low sparsity. 
SAPF \citep{fan2020sparse} also uses ADMM with projections to solve a factorized objective. It fails to converge for a tighter sparsity budget, requires extensive hyperparameter tuning, and has a prohibitively long running time.
Among generator-based methods, \cite{dong2020greedyfool} propose GreedyFool, a two-stage approach to greedily sparsify perturbations obtained from a generator.
Similarly, TSAA \citep{he2022transferable} generates sparse, magnitude-constrained adversarial attacks with high black-box transferability. The perturbations are typically spatially contiguous and are therefore more noticeable than other sparse attacks. The design of our attack, and employing Frank-Wolfe for optimization, yield highly sparse and inconspicuous adversarial examples efficiently.

\paragraph{Understanding Adversarial Attacks} 
A fairly novel research direction examines adversarial examples and model explanations in conjunction, noting the overlap between core ideas in the domains. On simple datasets such as MNIST, \cite{ignatiev2019relating} demonstrates a hitting set duality between model explanations and adversarial examples i.e. there exists a tight overlap between features identified as important for a model’s prediction and those targeted by successful adversarial attacks (and that one can be recovered from the other), suggesting that both explanations and attacks often focus on the same critical input features. Similarly, \citet{pmlr-v162-wang22e} leverage adversarial attacks to devise a novel model explainer. \citet{xu2019structured} examine the correspondence of attack perturbations with discriminative image features. We formulate our attack with an explicit sparsity constraint, which emphasizes only the most vulnerable pixels in an image. We also empirically analyze the overlap of adversarial perturbations and salient regions in images.

\section{Background}\label{sec:background}
In this section, we introduce the notations and conventions for adversarial attacks. We also provide a brief review of the Frank-Wolfe algorithm.

\subsection{Adversarial Attacks} Given an image $\mathbf{x} \in \mathbb{R}^{h\times w \times ch}$, a trained classifier $f : \mathbb{R}^{h\times w \times {ch}}\xrightarrow{}\{1\ldots C\}$ that maps the image to one of $C$ classes, and $f(\mathbf{x})=c$. Adversarial attacks aim at finding $\mathbf{x}'$ that is very similar to $\mathbf{x}$ by a distance metric, i.e. $||\mathbf{x}-\mathbf{x}'||_p \leq \epsilon$,  $(p \in \{0, 1, 2, \infty\},$ $\epsilon$ is small) such that $f(\mathbf{x}')=t$, where $t \neq c$.

Depending on the adversary's knowledge of the target model, adversarial attacks can be white-box (known model architecture and parameters) or black-box (unknown learning algorithm; the attacker only sees the most likely prediction given an input). Whether SAIF can be extended to the black-box setting (via, e.g., gradient approximations \citep{chen2020frank}) is not addressed in this work and deserves study.

\subsection{Frank Wolfe Algorithm}\label{sec:fw-bg}
The Frank-Wolfe algorithm (FW) \citep{frank1956algorithm} is a first-order, projection-free algorithm for optimizing a convex function $f(\mathbf{x})$ over a convex set $\mathbf{X}$. It is a projection-free method since it solves a linear approximation, known as the Linear Minimization Oracle (LMO), of the objective over $\mathbf{X}$. The key advantage of FW is that the iterates $\mathbf{x}_t$ always remain feasible $(\mathbf{x}_t \in \mathbf{X})$ throughout the optimization process. The algorithm was popularized for machine learning applications by \cite{jaggi2013revisiting} with rigorous proofs in objective value $f(\mathbf{x}_t) - f(\mathbf{x}^*)$, where $\mathbf{x}^*$ is the optimal point.

The first work using Frank-Wolfe for adversarial attacks \citep{chen2020frank} constrains only the magnitude of perturbation $\|\mathbf{x}-\mathbf{x}'\|_{\infty}$. As a result, the crafted attack is non-sparse. Later works also employ Frank-Wolfe for explaining predictions \citep{roberts2021controllably} and for faster adversarial training \citep{tsiligkaridis2022understanding, pmlr-v97-wang19i}.

Our motivation to employ FW for optimizing SAIF is twofold: \textbf{(1)} it is a `conservative' algorithm with iterates strictly in the feasible region throughout the optimization. Its projection-free nature prevents sub-optimal solutions common in methods like PGD, and \textbf{(2)} it has sparsity-inducing properties, which fits our goal.

\section{Method}\label{sec:method}
Our goal is to calculate an adversarial attack that has low magnitude and high sparsity simultaneously. Formally, the perturbations should have low $\ell_0$-norm and low $\ell_{\infty}$-norm to satisfy the sparsity and magnitude requirements, respectively. Moreover, the (untargeted) attack should maximize classification loss for the true class.

To implement such an attack, we define a sparsity-constrained mask $\mathbf{s}$ to preserve pixels of an additive adversarial perturbation $\mathbf{p}$. We also impose an $\ell_{\infty}$ constraint on the magnitude of $\mathbf{p}$. Decoupling the attack into a sparse mask and perturbation also allows visualizing the vulnerable pixels of the image.

\paragraph{Untargeted Attack.} Given $f(\mathbf{x})=c$, we define our untargeted objective function $D(\mathbf{s, p})_{adv}$ as
\begin{equation}\label{eqn:adv_D(s)}
    D(\mathbf{s,p})_{adv}= 
    \Phi(\mathbf{x} + \mathbf{s} \odot \mathbf{p}, c)
\end{equation}
Here $\Phi(.,c)$ is the classification loss function (e.g., cross-entropy) with respect to the true class $c$.

Note that optimization over the $\ell_0$ constraint is NP-hard. We use $\ell_1$ as the tightest convex approximation for $\ell_0$ over $\mathbf{s}$ following the common practice in the literature \citep{pmlr-v162-macdonald22a, he2022transferable}.
Thus, the optimization objective is to maximize the loss for the original class as:
\begin{equation}\label{eqn:adv_obj}
    \begin{aligned}
    \max _{\mathbf{s}, \mathbf{p}} D(\mathbf{s, p})_{adv}, \  \text {s.t.}\|\mathbf{s}\|_{1} \leq k, \ \mathbf{s} \in[0,1]^{h\times w\times ch},
    \|\mathbf{p}\|_\infty \leq \epsilon
    \end{aligned}
\end{equation}

This formulation not only highlights the vulnerable regions of the image to perturb via $\mathbf{s}$, but also yields an adversarial attack method where we can explicitly control the sparsity using $k$ and the perturbation magnitude per pixel with $\epsilon$. 

\paragraph{Targeted Attack.} We extend (\ref{eqn:adv_D(s)}) to targeted attacks by replacing $c$ with a chosen target class $\tilde{c}$.
\begin{equation}\label{eqn:D(s, p),t}
    \begin{aligned}
        D(\mathbf{s, p})_{\tilde{c}, adv} =  \Phi(\mathbf{x} + \mathbf{s} \odot \mathbf{p}, \tilde{c}), \quad \tilde{c} \neq c
    \end{aligned}
\end{equation}
Then to enhance the odds of predicting $\tilde{c}$, we minimize $D(\mathbf{s})_{\tilde{c},adv}$ to obtain the SAIF attack:
\begin{equation}\label{eqn:saif-obj}
    \begin{aligned}
    \min _{\mathbf{s}, \mathbf{p}} D(\mathbf{s, p})_{\tilde{c},adv}, \  \text {s.t.}\|\mathbf{s}\|_{1} \leq k, \ \mathbf{s} \in[0,1]^{h\times w\times ch}, 
    \|\mathbf{p}\|_\infty \leq \epsilon
    \end{aligned}
\end{equation}

\paragraph{Optimization.} We use Frank-Wolfe as the solver for our objectives \ref{eqn:adv_obj} and \ref{eqn:saif-obj} in order to ensure that the variable iterates remain feasible (see Algorithm \ref{alg:fw-saif}). The algorithm proceeds by moving the iterates towards a minimum by simultaneously minimizing the objective w.r.t. $\mathbf{s}$ and $\mathbf{p}$.

To constrain $\mathbf{s}$ we use a non-negative $k$-sparse polytope, which is a convex hull of the set of vectors in $[0, 1]^{h\times w\times \textcolor{red}{ch}}$, each vector admitting at most $k$ non-zero elements. We adopt the method in \cite{pmlr-v162-macdonald22a} to perform the LMO over this polytope.
That is, for $\mathbf{z_t}$ we choose the vector with at most $k$ non-zero entries, where the conditional gradient $\textbf{m}_t^s$ assumes $k$ smallest negative values (thus highest in magnitude). These $k$ components of $\mathbf{z_t}$ are then set to 1 and the rest to zero. For example, if $k=10$ and there are 20 negative values in $\textbf{m}_t^s$, the 10 smallest values are set to 1 and the rest to 0. 

For $\mathbf{p}$, the LMO of $\ell_\infty$ has a closed-form solution \citep{chen2020frank}.
\begin{align}
    \mathbf{v}_t = -\epsilon \cdot \text{sign}(\mathbf{m}^p_t) + \mathbf{x}
\end{align}

Note that it is possible to combine $\mathbf{s}$ and $\mathbf{p}$ into one variable using a method such as \cite{pmlr-v84-gidel18a}. However, in doing so we would lose the interpretability brought by disentangling the sparse mask $\mathbf{s}$. This is because enforcing an $\ell_1$ constraint is not the same as the currently enforced
$k$-sparse polytope. Therefore, such a dual constraint would result in a perturbation of varying values which is harder to interpret than a $[0,1]$-valued mask.

Since the objective of SAIF is non-convex, a monotonicity guarantee is helpful to ensure that the separate optimizations of each variable sync well.
Such monotonicity guarantee facilitates coordinated convergence of the sparse mask and perturbation variables, ensuring that both sparsity and magnitude constraints are jointly optimized in a stable and synchronized manner.
To this end, we use the following adaptive step size formulation \citep{carderera2021simple, pmlr-v162-macdonald22a} to ensure monotonicity in the objective
, which, together with the iterates being convex combinations, ensures that solutions do not leave the domain of feasible solutions:
\begin{align}
\eta_{t}=\frac{1}{2^{r_t}\sqrt{t+1}}
\end{align}
where we choose the $r_t\in\mathbb{N}$ by increasing from $r_{t-1}$, until we observe primal progress of the iterates. This method is conceptually similar to the backtracking line search technique often used with standard gradient descent. 

\begin{algorithm}
    \SetAlgoLined
    \SetKwInput{KwInput}{Input}
    \SetKwInput{KwOutput}{Output}
    \KwInput{Clean image $\mathbf{x} \in [0,I_{max}]^{h \times w \times c}$,
    $\textbf{s}_0 \in \mathcal{C}_s=\{\textbf{s} \in [0,1]^{h\times w\times ch} : \|\textbf{s}\|_1 \leq k\}$,
    $\textbf{p}_0 \in \mathcal{C}_p=\{\textbf{p} \in [0,I_{max}]^{h\times w\times ch}: \|\textbf{p}\|_\infty \leq \epsilon\}$.
    }
    \KwOutput{Perturbation $\textbf{p}$, Sparse mask $\textbf{s}$}
    \For{\text{t} $=1, \dots, T$}{
    $\textbf{m}^p_t = \nabla_p D(\textbf{s}_{t-1}, \textbf{p}_{t-1})$
    
    $\textbf{m}^s_t = \nabla_s D(\textbf{s}_{t-1}, \textbf{p}_{t-1})$
    
    
    
    $\textbf{v}_t = \text{argmin}_{\textbf{v} \in \mathcal{C}_p} \left<\textbf{m}^p_t, \textbf{v}\right>$
    
    $\textbf{z}_t = \text{argmin}_{\textbf{z} \in \mathcal{C}_s} \left<\textbf{m}^s_t, \textbf{z}\right>$
    
    $\textbf{p}_{t} = \textbf{p}_{t-1} + \eta_t(\textbf{v}_t - \textbf{p}_{t-1})$
    
    $\textbf{s}_{t} = \textbf{s}_{t-1} + \eta_t(\textbf{z}_t - \textbf{s}_{t-1})$
    
    }
\caption{\label{alg:fw-saif} SAIF - Adversarial attack using Frank-Wolfe for joint optimization.
}
\end{algorithm}

\section{Experiments}\label{sec:experiments}

We evaluate SAIF against several existing methods for both targeted and untargeted attacks. We report performance on the effectiveness as well as saliency of adversarial attacks.

\paragraph{Dataset and Models}
We use the ImageNet classification dataset (ILSVRC2012)~\citep{krizhevsky2012imagenet} in our experiments, which has $[299\times 299]$ RGB images belonging to 1,000 classes. We evaluate all attacks on 5,000 samples chosen from the validation set.
For classification, we test on two deep convolutional neural network architectures, namely Inception-v3 (top-1 accuracy: 77.9\%) and ResNet-50 (top-1 accuracy: 74.9\%). We use the pre-trained models from Keras applications \citep{kerasapps}.
We also report results on CIFAR-10 in the appendix.

\paragraph{Implementation}
We implement the experiments in Julia and use the Frank-Wolfe variants library \citep{besancon2021frankwolfejl}. We code the classifier and gradient computation backend in Python using TensorFlow and Keras deep learning frameworks. The experiments are run on a single Tesla V100 SXM2 GPU, for an empirically chosen number of iterations $T$ for each dataset. SAIF typically converges in $\sim$20 iterations, however, we relax the maximum iterations to $T=100$ in our experiments. 

\subsection{Evaluation Metrics}\label{sec:metrics}
\paragraph{Adversarial Attacks.} Adversarial attacks are commonly evaluated by the attack success rate.

\begin{itemize}
    \item \textbf{Attack Success Rate (ASR).}
A targeted adversarial attack is deemed successful if perturbing the image fools the classifier into labeling it with a premeditated target class $\tilde{c}$. An untargeted attack is successful if it leads the classifier to predict \emph{any} incorrect class. Given $n$ images in a dataset, if $m$ attacks are successful, the attack success rate is defined as ASR $=  m/n (\%)$.
\end{itemize}

Note that for RGB images, SparseFool \citep{modas2019sparsefool} and Greedyfool \citep{dong2020greedyfool} average the perturbation $\mathbf{p}$ across the channels and report the ASR for sparsity $||\mathbf{p}_{\text{flat}}||_0\leq k$, where $\mathbf{p}_{\text{flat}}\in [0,I_{max}]^{h\times w}$. Since for SAIF the sparsity constraint $k$ applies across all $h\times w\times \textcolor{red}{ch}$ pixels, we report the ASR for all methods \textbf{without} averaging the final perturbation across the channels (i.e. for $||\mathbf{p}||_0\leq k$, $\mathbf{p}\in [0,I_{max}]^{h\times w\times \textcolor{red}{ch}}$).

\begin{table}
\footnotesize
\begin{center}
\begin{tabular}{c | c || c | c | c | c | c || c | c | c | c | c}
    \toprule
    \multirow{2}{*}{$\|\mathbf{p}\|_{\infty}\leq \epsilon$} & \multirow{2}{*}{\bf Attacks} & \multicolumn{5}{c||}{\bf Inception-v3} & \multicolumn{5}{c}{\bf ResNet50}\\
    \cline{3-12}
    & & \multicolumn{5}{c||}{Sparsity `$k$'} & \multicolumn{5}{c}{Sparsity `$k$'} \\
    \hline
    \multicolumn{2}{c||}{ } & 100 & 200 & 600 & 1000 & 2000 & 100 & 200 & 600 & 1000 & 2000 \\
    \hline
    \multirow{4}{*}{$\epsilon$ = 255} & GreedyFool 
    & 0.40 & 1.19 & 19.36 & 49.70 & 87.82 & 0.59 & 2.59 & 28.94 & 70.06 & 96.21 \\
    & TSAA 
    & 0.00 & 1.15 & 31.61 & 77.01 & \textbf{100.0} &  -* & -* & -* & -* & -* \\
    & Homotopy-Attack 
    & 0.00 & 18.23 & 90.97 & \textbf{100.0} & \textbf{100.0} & 0.00 & 0.00 & 0.00 & 0.00 & 0.00 \\
    & SAIF (Ours)                           &  \textbf{55.97} & \textbf{84.00} & \textbf{100.0} & \textbf{100.0} & \textbf{100.0} &  \textbf{80.10} & \textbf{100.0} & \textbf{100.0} & \textbf{100.0} & \textbf{100.0} \\
    \hline
    \multicolumn{2}{c||}{ } & 100 & 200 & 600 & 1000 & 2000 & 100 & 200 & 600 & 1000 & 2000 \\
    \hline
    \multirow{3}{*}{$\epsilon$ = 100} & GreedyFool 
    & 0.20 & 0.40 & 4.59 & 12.18 & 25.35 & 0.20 & 1.59 & 12.57 & 28.54 & 44.11 \\
    & Homotopy-Attack 
    &  9.04 & 18.27 & 72.07 & \textbf{100.0} & \textbf{100.0} & 0.00 & 0.00 & 0.00 & 0.00 & 0.00 \\
    & SAIF (Ours)             &  \textbf{21.73} & \textbf{66.27} & \textbf{100.0} & \textbf{100.0} & \textbf{100.0} &  \textbf{59.01} & \textbf{90.72} & \textbf{100.0} & \textbf{100.0} & \textbf{100.0} \\
    \hline
    \multicolumn{2}{c||}{ } & 1000 & 2000 & 3000 & 4000 & 5000 & 1000 & 2000 & 3000 & 4000 & 5000 \\
    \hline
    \multirow{3}{*}{$\epsilon$ = 10} & GreedyFool 
    & 0.20 & 1.39 & 2.99 & 5.59 & 8.98 & 5.59 & 18.36 & 30.14 & 39.12 & 47.50 \\
    & Homotopy-Attack 
    & 0.00 & 30.05 & 58.27 & 69.59 & 85.03 & 0.00 & 0.00 & 0.00 & 0.00 & 0.00 \\
    & SAIF (Ours)                           &  \textbf{14.28} & \textbf{44.79} & \textbf{90.29} & \textbf{94.97} & \textbf{100.0} & \textbf{60.19} & \textbf{80.02} & \textbf{89.93} & \textbf{90.42} & \textbf{100.0}  \\
    \bottomrule
\end{tabular}
\end{center}
\caption{Quantitative evaluation of \textbf{targeted} attack on ImageNet. We report the ASR for varying constraints on sparsity `$k$' and $\ell_{\infty}$-norm of the magnitude of perturbation `$\epsilon$'. (* TSAA codebase lacks pre-trained generators for targeted attacks on ResNet50 and lower $\epsilon$.)}
\label{tab:main-results}
\end{table}

\begin{figure}[htp]
    \centering
    \includegraphics[width=0.98\linewidth]{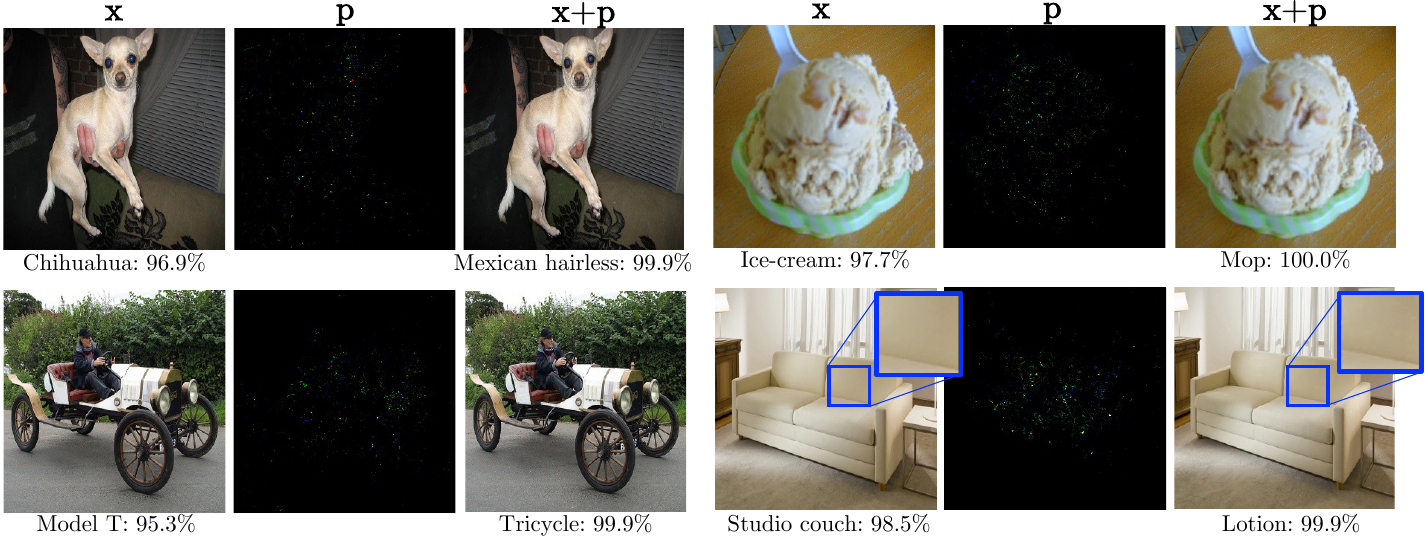}
    \caption{
    Qualitative results of \textbf{targeted} SAIF attack on Inception-v3 trained on ImageNet, using $\epsilon=100$ (39\% of the dynamic range of image) and $k=400$ (0.15\% of pixels). The source and target class, along with the corresponding probability, are stated below each $\mathbf{x}$ and $\mathbf{x+p}$ respectively.
    }
    \label{fig:perturb_samples}
\end{figure}

\paragraph{Attack Saliency.} We use the following metric to capture the correspondence between the vulnerable and salient pixels in the input images.
The score represents the overlap of the ground-truth (GT) bounding box of the subject of the input image with the (sparse) adversarial perturbation.

\begin{itemize}
    \item \textbf{Localization (Loc.) \citep{chattopadhay2018grad}} Effectively the same as IoU for object detection. Given image pixels X, GT salient pixels S 
    and SAIF sparse mask A, the localization score is:
\begin{equation}\label{eqn:loc_def}
    \text{Loc.} = \frac{\|A \cap S\|_0}{\|S\|_0+\|A \cap (X\setminus S)\|_0}
\end{equation}
    In the event of perfect correspondence between the GT salient regions and adversarial perturbation, Loc.$\rightarrow$1. Whereas, Loc.$\rightarrow$0 when there is poor overlap between the two.
\end{itemize}

\section{Results}
\subsection{Quantitative Results.}
 
\begin{table}[ht]
\footnotesize
\begin{center}
\begin{tabular}{c | c || c | c | c | c | c || c | c | c | c | c}
    \toprule
    \multirow{2}{*}{$\|\mathbf{p}\|_{\infty} \leq \epsilon$} & \multirow{2}{*}{\bf Attacks} & \multicolumn{5}{c||}{\bf Inception-v3} & \multicolumn{5}{c}{\bf ResNet50}\\
    \cline{3-12}
    & & \multicolumn{5}{c||}{Sparsity `$k$'} & \multicolumn{5}{c}{Sparsity `$k$'} \\
    \hline
    \multicolumn{2}{c||}{ } & 10 & 20 & 50 & 100 & 200 & 10 & 20 & 50 & 100 & 200 \\
    \hline
    \multirow{7}{*}{$\epsilon$ = 255} & SparseFool 
    & 1.59 & 4.39 & 15.37 & 32.14 & 32.14 & 1.79 & 3.19 & 8.78 & 17.76 & 33.33 \\
    & GreedyFool 
    & 3.99 & 7.58 & 16.57 & 35.33 & 62.87 & 4.19 & 7.78 & 21.76 & 42.12 & 72.26 \\
    & TSAA 
    &  0.00 & 0.00 & 0.00 & 0.00 & 2.02 &  0.00 & 0.00 & 0.00 & 1.95 & 16.06 \\
    & PGD $\ell_0+\ell_\infty$* 
    & 0.81 & 0.81 & 3.63 & 5.65 & 7.80 & 11.20 & 11.20 & 11.67 & 12.25 & 12.25 \\
    
    & PGD $\ell_{0}+\sigma$* 
    & 0.00 & 0.00 & 0.52 & 1.55 & 2.78 & 0.00 & 0.00 & 0.84 & 4.92 & 6.54 \\

    & SA-MOO* 
    &  9.52 & 10.04 & 14.28 & 38.09 & 39.47 &  27.98 & 44.32 & 45.12 & 54.83 & 60.91 \\
    & SAIF (Ours)             & \textbf{19.88} & \textbf{60.16} & \textbf{90.05} & \textbf{100.0} & \textbf{100.0} & \textbf{38.25} & \textbf{61.72} & \textbf{100.0} & \textbf{100.0} & \textbf{100.0} \\
    \hline
    \multicolumn{2}{c||}{ } & 10 & 20 & 50 & 100 & 200 & 10 & 20 & 50 & 100 & 200 \\
    \hline
    \multirow{6}{*}{$\epsilon$ = 100} & SparseFool 
    &  0.79 & 3.39 & 9.98 & 27.54 & 48.90 & 1.39 & 2.59 & 7.58 & 19.56 & 35.72 \\
    & GreedyFool 
    &  2.39 & 3.79 & 10.18 & 23.15 & 45.11 & 2.20 & 5.99 & 15.77 & 34.73 & 61.68 \\
    & PGD $\ell_0+\ell_\infty$* 
    & 0.00 & 0.24 & 3.29 & 5.22 & 6.86 & 11.67 & 12.25 & 12.25 & 12.25 & 14.49 \\

    & PGD $\ell_{0}+\sigma$* 
    & 0.00 & 0.00 & 0.28 & 0.62 & 2.49 & 0.00 & 0.00 & 0.00 & 3.78 & 5.92 \\
    
    & SA-MOO* 
    &  \textbf{4.67} & 9.52 & 9.98 & 14.28 & 23.81 &  \textbf{29.86} & 34.10 & 35.56 & 39.83 & 50.24 \\
    & SAIF (Ours)                           &  0.00 & \textbf{28.91} & \textbf{60.26} & \textbf{90.03} & \textbf{100.0} & 20.42 & \textbf{41.26} & \textbf{79.32} & \textbf{100.0} & \textbf{100.0} \\
    \hline
    \multicolumn{2}{c||}{ } & 200 & 500 & 1000 & 2000 & 3000 & 200 & 500 & 1000 & 2000 & 3000 \\
    \hline
    \multirow{7}{*}{$\epsilon$ = 10} & SparseFool 
    & 4.19 & 14.17 & 38.92 & 67.86 & 82.24 & 11.98 & 39.92 & 69.46 & 90.02 & 95.61 \\
    & GreedyFool 
    &  8.78 & 22.55 & 40.52 & 65.07 & 77.25 & 18.77 & 47.70 & 74.65 & 93.41 & 97.21 \\
    & TSAA 
    &  0.00 & 0.00 & 0.00 & 0.00 & 0.00 & 0.00 & 0.00 & 0.00 & 0.59 & 6.89 \\
    & PGD $\ell_0+\ell_\infty$* 
    & 4.83 & 7.69 & 11.21 & 22.03 & 30.47 & 11.28 & 11.28 & 11.64 & 12.80 & 14.83 \\

    & PGD $\ell_{0}+\sigma$* 
    & 0.72 & 4.02 & 7.06 & 12.83 & 13.76 & 0.00 & 0.00 & 0.04 & 0.22 & 6.24 \\
    
    & SA-MOO* 
    &  4.76 & 4.98 & 5.02 & 9.52 & 10.98 &  15.56 & 17.82 & 18.01 & 18.94 & 20.97 \\
    & SAIF (Ours)                           &  \textbf{10.21} & \textbf{52.40} & \textbf{89.02} & \textbf{90.00} & \textbf{100.0} &  \textbf{50.04} & \textbf{73.09} & \textbf{95.00} & \textbf{100.0} & \textbf{100.0}  \\
    \bottomrule
\end{tabular}
\end{center}
\caption{Quantitative evaluation of \textbf{untargeted} attack on Inception-v3 and ResNet50 trained on ImageNet dataset. We report the ASR for varying constraints on sparsity `$k$' and $\ell_{\infty}$-norm of the magnitude of perturbation `$\epsilon$'. Black-box attacks are marked with *.}
\label{tab:main-results-2}
\end{table}

We evaluate the ASR of all attack methods on a range of constraints on perturbation magnitude $\epsilon$ and sparsity $k$. The results for \emph{targeted} attacks on ImageNet are reported in Table \ref{tab:main-results}. The target class is randomly chosen for each sample. Note that TSAA \citep{he2022transferable} does not evaluate attacks for $\epsilon=100$. Moreover, for both targeted and untargeted attacks, SAPF \citep{fan2020sparse} fails for all the evaluated thresholds.

Table \ref{tab:main-results} demonstrates that SAIF consistently outperforms both targeted attack baselines by a large margin for all $\epsilon$ and $k$ thresholds. For smaller $\epsilon$, GreedyFool fails to attack samples unless the sparsity threshold is significantly relaxed. A similar pattern is observed for the sparsity budget - competing attacks completely fail for tighter bounds on perturbation magnitude (see $\epsilon\in\{10,100,255\}$ at $k=2000$). Moreover, at lower $k$, ResNet50 is easier to fool than Inception-v3.

We also compare the ASR for \emph{untargeted} attacks against the baselines in Table \ref{tab:main-results-2}. Here as well SAIF significantly outperforms other attacks, particularly on tighter perturbation bounds. By jointly optimizing over the two constraints, we are able to fool DNNs with extremely small distortions. For instance, at $\epsilon=10$, SAIF modifies only 0.37\% pixels per image to successfully perturb 89.02\% of input samples. This is more than twice the ASR of state-of-the-art sparse attack methods. Similarly, at $\epsilon=255$, only 0.03\% pixels are attacked to achieve a perfect ASR. On CIFAR-10, SAIF achieves $\sim$3$\times$ higher ASR on lower sparsity thresholds. The results are reported in the Appendix.

\begin{figure}[ht]
\centering
\includegraphics[width=0.99\textwidth]
    {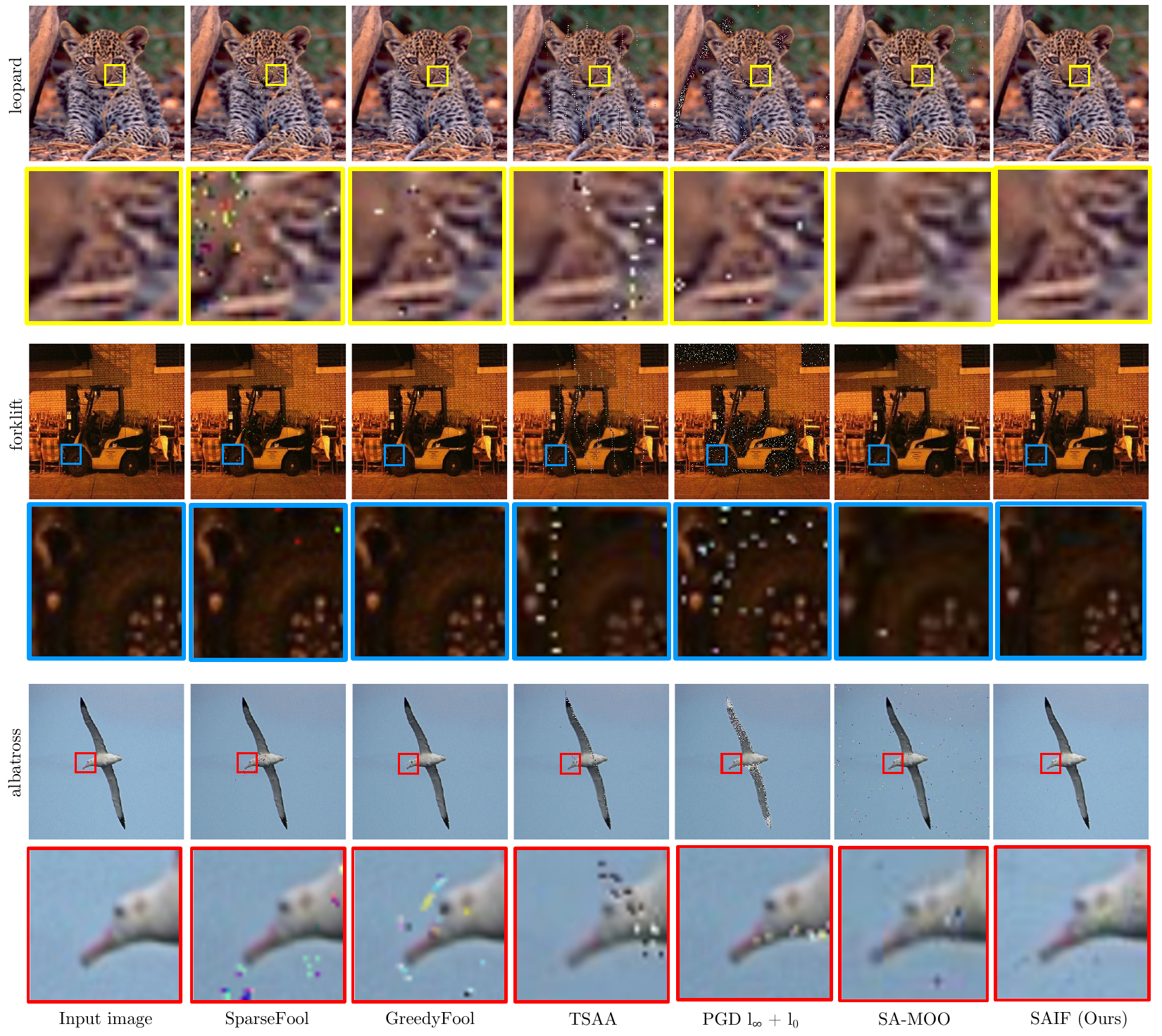}
    \caption{Visual results of \textbf{untargeted} attack on three images for $\epsilon=255$, $k=600$. Our method produces the most imperceptible adversarial examples despite the relaxed constraint on magnitude.}
    \label{fig:qual-comparison}
\end{figure}

Note that sparse attacks like ours address a more challenging problem than $\ell_\infty$-norm threat models since both the perturbation magnitude \textbf{and} sparsity are to be constrained. For $\ell_\infty$ constraints $\epsilon=$10, 100, and 255, AutoAttack \citep{croce2020reliable} achieves $100\%$ ASR while perturbing $\sim$98\% of image pixels for each $\epsilon$. For the same $\epsilon$, SAIF achieves $100\%$ ASR with \textbf{1.12\%, 0.07\%, and 0.037\%} sparsity, respectively. Moreover, sparse attacks are more visually interpretable than $\ell_\infty$ attacks.

We additionally evaluate SAIF against the One Pixel Attack \citep{su2019one}, which aims to fool classifiers by modifying only a single pixel per image. For ResNet50 trained on ImageNet, the One Pixel Attack is reported to achieve 34.0\% ASR for untargeted and 14.4\% ASR for targeted settings. Despite not being specifically optimized for such an extreme sparsity regime, SAIF achieves a fairly competitive 20.93\% ASR in the untargeted setting and 6.04\% in the targeted case, demonstrating its robustness even under highly constrained attack scenarios.

\subsection{Qualitative Results.}
We include some examples of targeted adversarial examples using SAIF in Figure \ref{fig:perturb_samples}. Note that the perturbations $\mathbf{p}$ have been enhanced in all figures for visibility. Visually, the perturbations generated by SAIF are only slightly noticeable in regions with a uniform color palette/low textural detail, such as on the beige couch (see the zoomed-in segments of Figure \ref{fig:perturb_samples}). The other images are bereft of such regions and thus have negligible visible change. Moreover, the attack predominantly perturbs semantically meaningful pixels in the images.

For untargeted attacks, we present examples from all competing attack methods in Figure \ref{fig:qual-comparison}. We ease the magnitude constraint to allow all baselines to achieve some successful attacks at the same level of sparsity.
Upon a closer visual inspection, it can be observed that competing methods produce significantly more conspicuous perturbations
around the face of the leopard and on the outlines of the forklift and the albatross. PGD $\ell_\infty + \ell_0$, in particular, adds the most noticeable distortions to images. In contrast, SAIF attack produces adversarial examples that appear virtually identical to the clean input images.

\subsection{Perturbations and Interpretability}\label{sec:interpretability}

\begin{table}
\begin{center}
\begin{tabular}{l | c} 
    \toprule
    \bf Attack           & \bf Loc. $\uparrow$ \\ 
    \hline
    SparseFool 
    &   0.006 \\ 
    GreedyFool 
    &   0.001 \\ 
    TSAA 
    &   0.006 \\ 
    \midrule
    SAIF (untargeted)  &   \textbf{0.126} \\ 
    SAIF (targeted)    &   0.118 \\ 
    \bottomrule
\end{tabular}
\end{center}
\caption{Quantitative evaluation of interpretability w.r.t. overlap with GT bounding boxes of ImageNet \textit{val} set. We use $\epsilon=255, k=200$ for all untargeted attacks unless indicated otherwise.}
\label{tab:saliency-results}
\end{table}

The sparse nature of perturbations allows us to study the interpretability of each attack (i.e. their correspondence with discriminative image regions). A similar evaluation is carried out in \cite{xu2019structured}, but they treat the saliency maps obtained from CAM \citep{zhou2016learning} as the ground truth. The saliency maps from CAM (and other existing methods) incur several failure cases. Therefore, we use the ImageNet bounding box annotations as a reliable baseline to analyze attack understandability.

We use ResNet50 as the target model and set $\epsilon=255$ and $k=2000$ for all attacks. The results are reported in Table \ref{tab:saliency-results}. SAIF achieves the highest Loc. score among competing methods in both the targeted and untargeted attack setting. Note that SAIF performs better on this metric in the untargeted attack setting versus the targeted attack. This is intuitive since the untargeted objective only diminishes features of the true class. Whereas, targeted attacks introduce features to convince a DNN to predict the target class.

\subsection{Comparison against non-sparse methods}
To demonstrate the importance of sparsity in adversarial attacks, we present a quantitative and qualitative comparison against non-sparse attack methods.
\paragraph{Percentage of pixels perturbed:} 
To emphasize the significance of an explicit sparsity constraint for adversarial attacks,
we report the percentage of pixels perturbed by three non-sparse attacks against SAIF. Comprehensive results are in \cref{tab:sparsities}. SAIF achieves perfect ASR by modifying $\leq$1\% of pixels. The same ASR is achieved by non-sparse baselines by modifying $\sim99\%$ pixels for all constraints on perturbation magnitude.

\begin{table}[ht]
    \centering
    \begin{tabular}{c|c|c|c|c}
    \toprule
        $\epsilon$ & FGSM  & PGD & AutoAttack  & SAIF \\
        \hline
        255 & 99.28 & 99.87 & 98.85 & \textbf{0.04} \\
        100 & 99.29 & 99.98 & 98.99 & \textbf{0.07} \\
        10 & 99.28 & 99.98 & 99.93 & \textbf{1.12} \\
    \bottomrule
    \end{tabular}
    \caption{Percentage of pixels attacked for 100\% ASR on ResNet50. $\epsilon$ is the constraint on perturbation magnitude.}
    \label{tab:sparsities}
\end{table}
    
\begin{figure}
    \centering
    \includegraphics[width=0.75\linewidth]{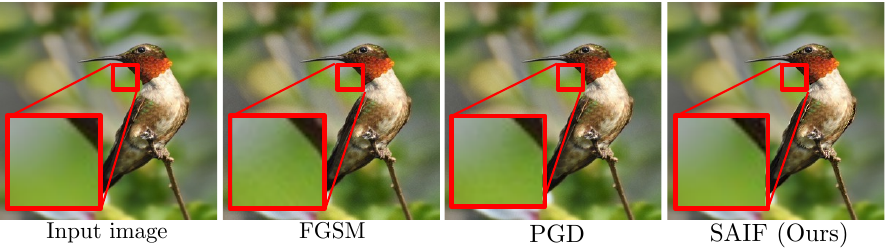}
    \vspace{-3mm}
    \caption{Attacks with $\epsilon$ = 2/255 on ResNet-50. (Best viewed at higher brightness levels).}
    \label{fig:imperceptibilty}
\end{figure}

\paragraph{Visual imperceptibility:} We evaluate all methods for $\epsilon$ = 2/255, which is an extremely low perturbation magnitude, and present the qualitative results in Figure \ref{fig:imperceptibilty}. Despite the very small $\epsilon$, non-sparse attacks are more noticeable than SAIF due to a lack of sparsity in the perturbations.

\subsection{Speed Comparison}
\begin{table}
\footnotesize
\centering
    \begin{tabular}{l|c}
        \toprule
        \textbf{Attack} & \textbf{Time (sec)} \\
        \midrule
         SparseFool 
         &  20 \\
         GreedyFool 
         &  1.7 \\
         TSAA 
         &  1.8 \\
         SAPF 
         &  1142 \\
         Homotopy-Attack 
         &  1500 \\
         PGD $\ell_0+\ell_\infty$ 
         &  \textbf{1.46} \\
         SA-MOO & 56.07 \\
         \hline
         SAIF (ours)                           &  15 \\
         \bottomrule
    \end{tabular}
    \caption{Average running time (per image) on ImageNet}
    \label{tab:runtimes}
\end{table}

Table \ref{tab:runtimes} reports running times of all baselines.
PGD $\ell_0+\ell_\infty$ \citep{croce2019sparse} runs the fastest but produces the most noticeable perturbations. 
Note that, although the inference time for GreedyFool \citep{dong2020greedyfool} and TSAA \citep{he2022transferable} is $\leq 2$ seconds, these generator-based attacks require pre-training a generator for each target model (as well as each $\epsilon$ and target class for TSAA \citep{he2022transferable}). This incurs a significant computational overhead ($>$7 days on a single GPU), which offsets their faster optimization times. SAIF attack relies solely on pre-trained classifiers, and is significantly faster than the existing state-of-the-art Homotopy-Attack \citep{zhu2021sparse}. Moreover, more efficient implementations of the FW LMO can further shorten running times, which we leave for future work.

\section{Ablation Studies}\label{sec:ablation}
We perform two sets of ablative experiments to highlight the significance of our design choices.

\paragraph{Attack Sparsity.} To illustrate the importance of limiting the sparsity of attack using $\mathbf{s}$, we reformulate the problem to one constrained only over the perturbation magnitude. That is, we use the following objective for the untargeted attack:

\begin{equation}\label{eqn:adv_D(p)_untar}
    D(\mathbf{p})_{adv}= 
    \Phi(\mathbf{x} + \mathbf{p}, c)
\end{equation}
\begin{equation}\label{eqn:adv_untarg_obj_p_only}
    \begin{aligned}
    \max _{\mathbf{p}} D(\mathbf{p})_{adv}, \  \text { s.t. } \quad \|\mathbf{p}\|_\infty \leq \epsilon
    \end{aligned}
\end{equation}

Similarly, we reframe the targeted attack by optimizing for the objective $D(\mathbf{p})_{adv}$ where,
\begin{equation}\label{eqn:adv_D(p)_tar}
    D(\mathbf{p})_{t,adv}= 
    \Phi(\mathbf{x} + \mathbf{p}, t)
\end{equation}
\begin{equation}\label{eqn:adv_targ_obj_p_only-obj}
    \begin{aligned}
    \min _{\mathbf{p}} D(\mathbf{p})_{t,adv}, \  \text { s.t. } \quad \|\mathbf{p}\|_\infty \leq \epsilon
    \end{aligned}
\end{equation}

This is similar to \citet{chen2020frank}'s attack method that uses Frank-Wolfe for optimization. 

Following Table \ref{tab:main-results}-\ref{tab:main-results-2}, we test the attack for various $\epsilon$ and report the results in Table \ref{tab:abl-results}. In the absence of a sparsity constraint, the attack distorts all image pixels regardless of the constraint on magnitude. Moreover, such spatially `contiguous' perturbations are visible even at magnitudes as low as $\epsilon=10$ (see Figure 4). By constraining the sparsity for SAIF attack, we ensure the adversarial perturbations stay imperceptible even at $\epsilon=255$, at which the non-sparse attack completely obfuscates the image.

\begin{figure}
    \centering
    \includegraphics[width=0.6\linewidth]{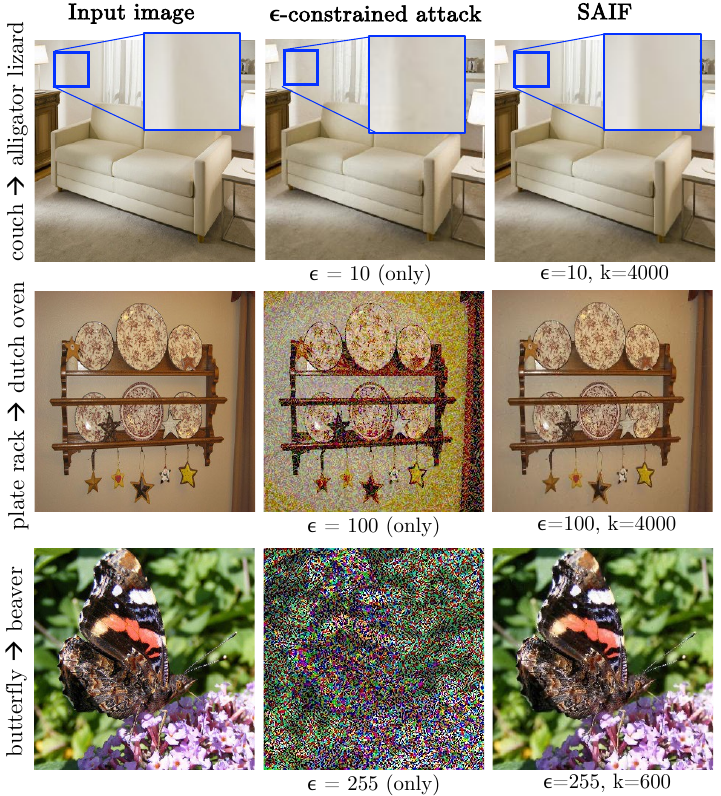}
    \caption{Exploring the significance of sparsity of the adversarial attack. When the sparsity is not constrained (middle column), perturbations of very small magnitude ($\epsilon = 10$) are noticeable and completely distort the image for larger $\epsilon$. On the contrary, SAIF (third column) stays imperceptible at higher magnitudes as well.}
    \label{fig:ablation-comp}
\end{figure}

\begin{table}
\small
\begin{center}
\begin{tabular}{c|c|c|c}
\toprule
    \bf Attack Type & \bf Dataset  & \bf ASR $\uparrow$ & $||\mathbf{p}||_{0}/m$ $\downarrow$ \\
    \hline
    \multirow{3}{*}{untargeted} & $\epsilon=255$ & 100.0 & 1.0 \\
    & $\epsilon=100$  & 100.0 & 1.0 \\
    & $\epsilon=10$   & 100.0 & 1.0 \\
    \hline
    \multirow{3}{*}{targeted} & $\epsilon=255$    & 100.0 & 1.0 \\
    & $\epsilon=100$  & 100.0 & 1.0 \\
    & $\epsilon=10$   & 100.0 & 1.0 \\
    \bottomrule
\end{tabular}
\end{center}
\caption{Quantitative evaluation of optimizing SAIF without a sparsity constraint for Inception-v3 on ImageNet. For each perturbation magnitude, the attack distorts all $m$ pixels in the image, leading to high perceptibility.}
\label{tab:abl-results}
\end{table}

\begin{table}[htp]
\
    \centering
    \begin{tabular}{l|c}
    \toprule
         \textbf{Constraints} & \textbf{ASR}  \\
         \hline
         $\epsilon=255, k=600$ & 97.78\% \\
         $\epsilon=100, k=600$ & 98.30\% \\
         $\epsilon=10, k=3000$ & 88.97\% \\
     \bottomrule
    \end{tabular}
    \caption{ASR for targeted attacks on Inception-v3 when cross-entropy is replaced with $\ell_2$-attack loss~\citep{carlini2017towards}.}
    \label{tab:loss_formula}
\end{table}

\paragraph{Loss formulation.} We also experiment with different losses, mainly the $\ell_2$-attack proposed by \cite{carlini2017towards}, but observe a decline in attack success (see Table \ref{tab:loss_formula} - we choose $\epsilon$ and $k$ for which SAIF achieves 100\% ASR in Table \ref{tab:main-results}.). A possible explanation for this behavior is that the loss formulation tries to increase the target class probability too aggressively, which makes the simultaneous optimization of $\mathbf{s}$ and $\mathbf{p}$ difficult. We observe that this yields solutions closer to the constraint boundaries, increasing the attack visibility.

\section{Conclusion}\label{sec:conclusion}
In this work, we propose a novel adversarial attack, `SAIF', by jointly minimizing the magnitude and sparsity of perturbations. By constraining the attack sparsity, we not only conceal the attacks but also identify the most vulnerable pixels in natural images. We use the Frank-Wolfe algorithm to optimize our objective and achieve effective convergence, with reasonable efficiency, for large natural images. We perform comprehensive experiments against state-of-the-art attack methods and demonstrate the remarkably superior performance of SAIF under tight magnitude and sparsity budgets. Our method also outperforms existing methods on a quantitative metric for interpretability and provides transparency to visualize the vulnerabilities of DNNs.

\section*{Discussion and Ethics Statement}\label{sec:ethics}
Adversarial attacks expose the fragility of DNNs. Our work aims to demonstrate that a highly imperceptible adversarial attack can be generated for natural images. This provides a new benchmark for the research community to test the robustness of the learning algorithms. A straightforward defense strategy can be using the adversarial examples generated by SAIF for adversarial training. We leave more advanced solutions for future exploration.

Moreover, in principle, SAIF can be extended to other input modalities, with appropriate adaptations for the target domain. For example, extending SAIF to video data would require adding a temporal consistency constraint, alongside the existing sparsity and magnitude constraints, to preserve the imperceptibility of perturbations. We leave a detailed exploration of these potential extensions to future work.

\section*{Acknowledgments}\label{sec:ack}
This project was supported by NIH/NCI grant R01CA240771.

\bibliography{saif}
\bibliographystyle{tmlr}

\appendix
\section*{Appendix}
\section{Frank Wolfe Algorithm}
The Frank-Wolfe algorithm (FW) \citep{frank1956algorithm} is a first-order, projection-free algorithm for optimizing a convex function $f(\mathbf{x})$ over a convex set $\mathbf{X}$ (Algorithm \ref{alg:FW}).

The set $\mathbf{X}$ may be described as the convex hull of a (possibly infinite) set of atoms $\mathcal{A}$. 
In the case of the $\ell_1$ ball $(\mathbf{X} = \{\mathbf{x} \in \mathbb{R}^n \mid \|\mathbf{x}\|_1 \leq \tau\})$, 
these atoms may be chosen as the $2n$ unit vectors, i.e., $\mathcal{A} = \{\pm  \mathbf{e}_j, \ j=1\ldots n\}$.

\begin{algorithm}
    \caption{\label{alg:FW} Frank-Wolfe Algorithm}
    \SetAlgoLined
    \SetKwInput{KwInput}{Input}
    \SetKwInput{KwOutput}{Output}
    \KwInput{Objective $f$, convex set $\mathcal{X}$, Maximum iterations $T$, stepsize rule $\eta_t$}
    \KwOutput{Final iterate $\mathbf{x_T}$}
 $\mathbf{x_0} = \mathbf{0}$
 
    \For{$t=0\ldots T-1$}{
        $\mathbf{a_t}=\argmin_{\mathbf{a} \in \mathcal{X}} \left<\nabla f(\mathbf{x_t}),\mathbf{a}\right>$ 
        
         $\mathbf{x_{t+1}} = \mathbf{x_t} + \eta_t (\mathbf{a_t} - \mathbf{x_t})$
    }
\end{algorithm}

FW is a projection-free method since it solves a linear approximation of the objective over $\mathbf{X}$ (see step 3 in Algorithm \ref{alg:FW}), known as the Linear Minimization Oracle (LMO). 
For convex optimization, the optimality gap $f(\mathbf{x}_t) - f(\mathbf{x}^*)$ is upper bounded by the duality gap $g(\mathbf{x}_t) = \min_{a \in \mathcal{A}} \left< \nabla f(\mathbf{x}_t), a\right>$, which measures the instantaneous expected decrease in the objective and converges at a sub-linear rate of $ O(1/T)$ for Algorithm \ref{alg:FW} \citep{jaggi2013revisiting}. FW can locally solve non-convex objectives over convex regions with $O(1/\sqrt{T})$ convergence
\citep{lacoste2016convergence}. 

\section{Results on Vision Transformers}

We also evaluate SAIF and GreedyFool (the best performing/reasonably efficient baseline) on a pre-trained ViT-B/16 \citep{dosovitskiy2021an} (see Table \ref{tab:tar-vit}). Although GreedyFool performs similarly, SAIF produces more visually imperceptible perturbations.

\begin{table}
\footnotesize
    \centering
    \begin{tabular}{c| c |c|c|c|c|c}
    \toprule
       \multirow{2}{*}{\textbf{$\epsilon$}} & \multirow{2}{*}{ \textbf{Attack}} & \multicolumn{5}{c}{\textbf{Sparsity `$k$'}} \\
       \cline{3-7}
       &  & 100 & 200 & 600 & 1000 & 2000 \\
    \hline
    \multirow{2}{*}{255} & GreedyFool [11] & 5.7 & 29.1 & 87.4 & \textbf{99.2} & 99.8 \\
    & SAIF & \textbf{28.3} & \textbf{54.4} & \textbf{89.4} & 93.5 & \textbf{100.0} \\
    \hline
    \multirow{2}{*}{100} & GreedyFool [11] & 1.7 & 14.3 & 54.5 & \textbf{82.0} & 92.4 \\
    & SAIF & \textbf{4.5} & \textbf{26.7} & \textbf{76.8} & 80.7 & \textbf{100.0} \\
    \hline
    \multicolumn{2}{c|}{} & 1000 & 2000 & 3000 & 4000 & 5000 \\
    \hline
    \multirow{2}{*}{10}  & GreedyFool [11] & \textbf{25.34} & \textbf{66.7} & \textbf{83.2} & \textbf{91.8} & \textbf{95.7} \\
    & SAIF & 2.9 & 22.4 & 34.9 & 42.0 & 52.3 \\
     \bottomrule
    \end{tabular}
    \caption{Targeted ASR (\textbf{higher is better}) on ViT-B/16 (Imagenet).}
    \label{tab:tar-vit}
\end{table}

\section{Results on additional dataset}
For a comprehensive evaluation of our attacks, we also test our approach on a smaller dataset:

\subsection{Dataset and model}
We test SAIF and the existing sparse attack algorithms on the CIFAR-10 dataset \citep{krizhevsky2009learning}. The dataset comprises $[32\times 32]$ RGB images belonging to 10 classes. We evaluate all algorithms on 10,000 samples from the test set.

We attack VGG-16 \citep{vgg-cifar} trained on CIFAR-10, having an accuracy of 92.32\% on clean images.

\subsection{Quantitative Results}
We run all attacks for a range of $\epsilon$ and $k$, and report the results in Tables \ref{tab:results-cifar},\ref{tab:results-cifar-tar}. 

The ASR for untargeted attack is reported in Table \ref{tab:results-cifar}. SAIF consistently outperforms SparseFool \citep{modas2019sparsefool} and GreedyFool \citep{dong2020greedyfool} on all sparsity and magnitude constraints. Similar results are obtained for targeted attacks, reported in Table \ref{tab:results-cifar-tar}. The target classes are randomly chosen in all experiments.

\subsection{Qualitative Results}
We provide several examples of adversarial examples produced by SAIF and competing algorithms. Figure \ref{fig:cifar-qual-untar} shows samples obtained by untargeted attacks on VGG-16 trained on CIFAR-10. It is observed that SAIF consistently produces the most imperceptible perturbations.

\section{Visual interpretability of SAIF}
\begin{figure}
    \centering
    \includegraphics[width=0.98\linewidth]{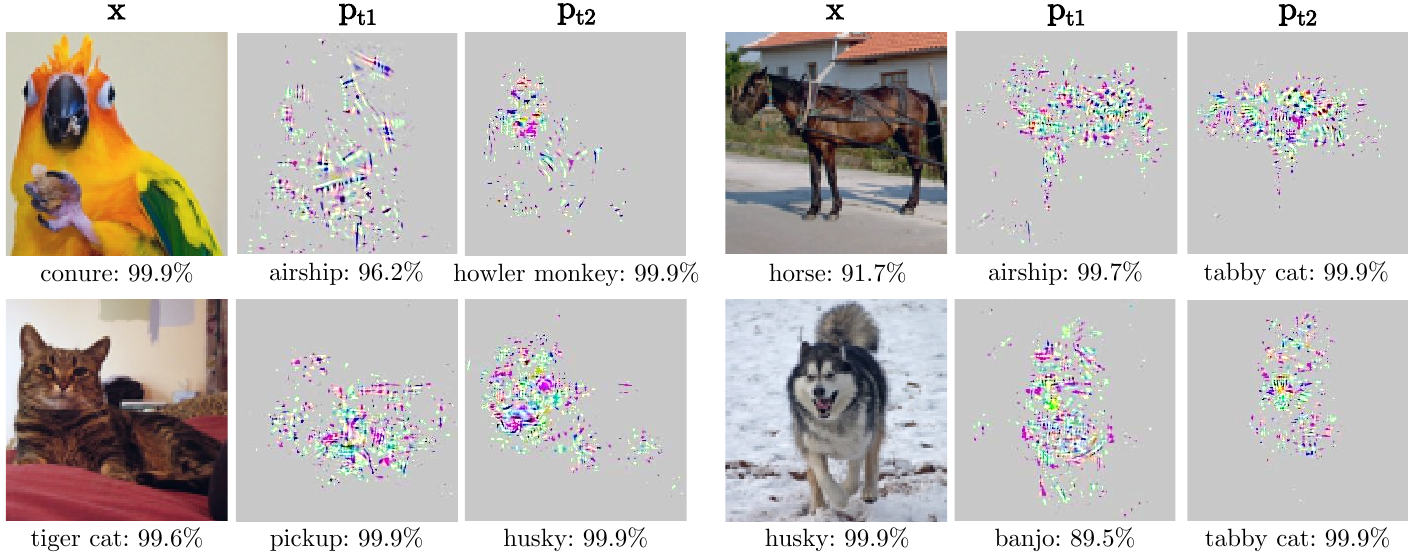}
    \caption{Impact of choice of target class on adversarial perturbations. We run targeted attacks for two target classes ($\mathbf{t_1}$ and $\mathbf{t_2}$) for each input. The perturbations are enhanced for visualization.
    }
    \label{fig:img_saliency}
\end{figure}

\begin{figure*}
    \centering
    \includegraphics[width=0.95\textwidth]{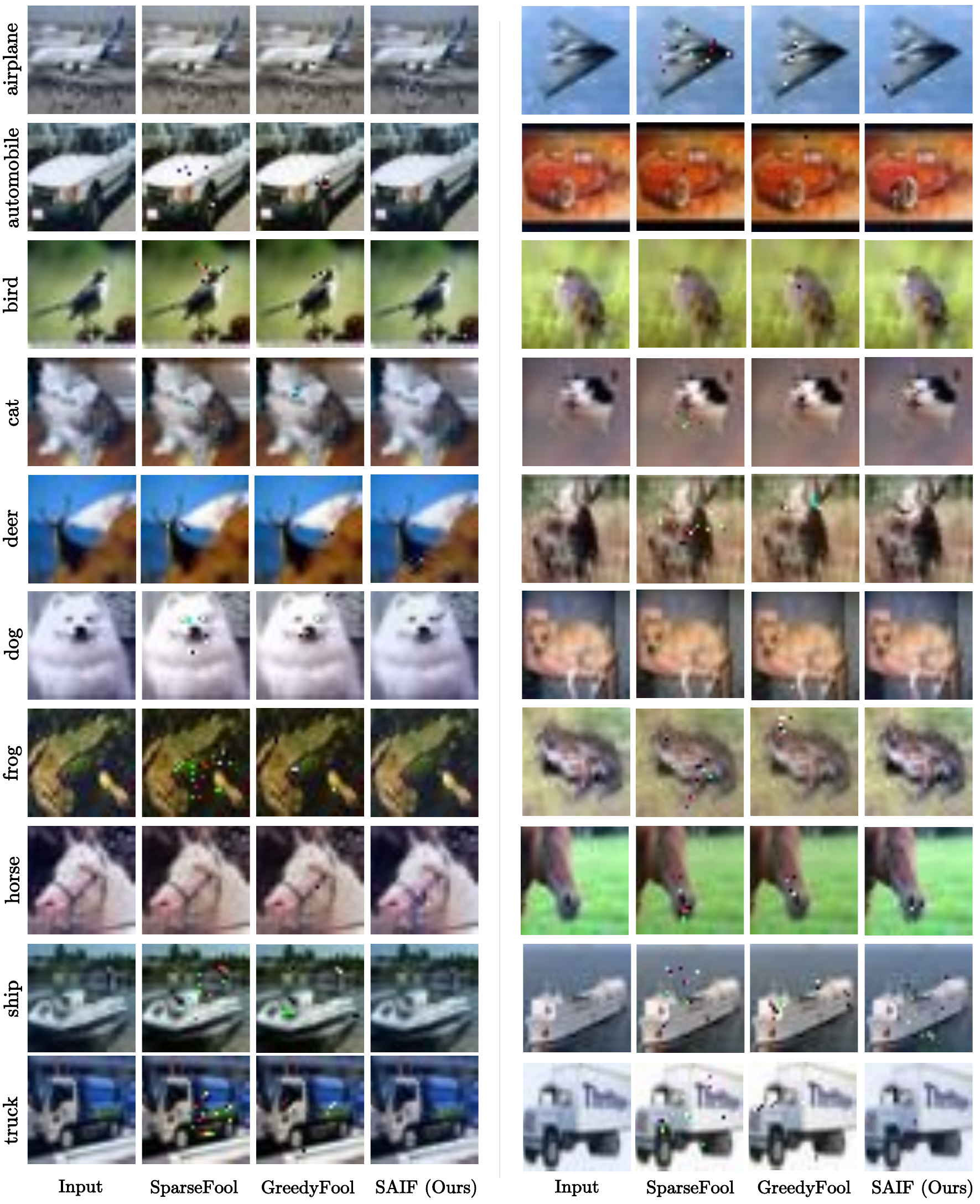}
    \caption{Untargeted adversarial examples obtained from SAIF and competing attack algorithms on CIFAR-10. The attacked classifier is VGG-16}
    \label{fig:cifar-qual-untar}
\end{figure*}

From visual inspection (see Figure \ref{fig:img_saliency}) it is observed that the nature of the target class also determines the sparse distortion pattern. In particular, attacking input images of an animate class towards another animate class ($t_2$) results in perturbations focused predominantly on the facial region in the image. The reverse is observed when attacking animate towards inanimate object classes ($t_1$), which typically modify the body of the subject in the image.

\begin{table*}[htp]
\footnotesize
\begin{center}
\begin{tabular}{c | c || c | c | c | c | c } 
    \toprule
    \multirow{2}{*}{$\|\mathbf{p}\|_{\infty}\leq \epsilon$} & \multirow{2}{*}{\bf Attacks} & \multicolumn{5}{c}{\bf VGG-16}\\ 
    \cline{3-7}
    & & \multicolumn{5}{c}{Sparsity `$k$'} \\ 
    \hline
    \multicolumn{2}{c}{ }           & 1 & 2 & 5 & 10 & 20 \\ 
    \hline
    \multirow{3}{*}{$\epsilon$ = 255} & SparseFool    & 10.78 & 18.56 & 38.32 & 63.67 & 85.23 \\ 
    & GreedyFool  &  0.00 & 0.00 & 24.75 & 69.26 & 85.83 \\ 
    & SAIF (Ours)                           &  \textbf{91.20} & \textbf{92.87} & \textbf{94.46} & \textbf{96.35} & \textbf{100.0} \\ 
    \hline
    \multicolumn{2}{c||}{ } & 5 & 10 & 15 & 20 & 30 \\ 
    \hline
    \multirow{3}{*}{$\epsilon$ = 100} & SparseFool  &  30.54 & 51.50 & 65.47 & 74.45 & 86.43 \\ 
    & GreedyFool &  25.95 & 55.09 & 67.26 & 73.65 & 86.43 \\ 
    & SAIF (Ours)                           &  \textbf{91.89} & \textbf{92.84} & \textbf{93.57} & \textbf{96.74} & \textbf{97.80} \\ 
    \hline
    \multicolumn{2}{c||}{ } & 30 & 40 & 50 & 60 & 100 \\ 
    \hline
    \multirow{3}{*}{$\epsilon$ = 10} & SparseFool    & 19.36 & 24.75 & 27.94 & 31.94 & 44.51 \\ 
    & GreedyFool  &  33.93 & 39.92 & 45.91 & 51.10 & 66.27 \\ 
    & SAIF (Ours)                           &  \textbf{90.32} & \textbf{91.57} & \textbf{92.14} & \textbf{92.65} & \textbf{94.14} \\ 
    \bottomrule
\end{tabular}
\end{center}
\vspace{-0.4cm}
\caption{Quantitative evaluation of \textbf{untargeted} attack on CIFAR-10. We report the ASR for varying constraints on sparsity `$k$' and $\ell_{\infty}$-norm of the magnitude of perturbation `$\epsilon$'.}
\label{tab:results-cifar}
\end{table*}

\begin{table*}[htp]
\footnotesize
\begin{center}
\begin{tabular}{c | c || c | c | c | c | c } 
    \toprule
    \multirow{2}{*}{$\|\mathbf{p}\|_{\infty}\leq \epsilon$} & \multirow{2}{*}{\bf Attacks} & \multicolumn{5}{c}{\bf VGG-16} \\ 
    \cline{3-7}
    & & \multicolumn{5}{c}{Sparsity `$k$'} \\
    \hline
    \multicolumn{2}{c}{ }           & 1 & 2 & 5 & 10 & 20 \\
    \hline
    \multirow{2}{*}{$\epsilon$ = 255} & GreedyFool  &  0.00 & 0.00 & 2.79 & 13.17 & 29.94 \\
    & SAIF (Ours)                           &  \textbf{12.36} & \textbf{12.83} & \textbf{27.02} & \textbf{44.05} & \textbf{61.10} \\
    \hline
    \multicolumn{2}{c||}{ } & 5 & 10 & 15 & 20 & 30 \\
    \hline
    \multirow{2}{*}{$\epsilon$ = 100} & GreedyFool &  2.20 & 9.58 & 14.97 & 16.97 & 30.74 \\
    & SAIF (Ours)                           &  \textbf{13.37} & \textbf{21.03} & \textbf{26.27} & \textbf{36.18} & \textbf{51.43} \\
    \hline
    \multicolumn{2}{c}{ } & 30 & 40 & 50 & 60 & 100 \\
    \hline
    \multirow{2}{*}{$\epsilon$ = 10} & GreedyFool & 3.99 & 5.19 & 6.99 & 8.98 & 16.17 \\
    & SAIF (Ours)                           &  \textbf{5.46} & \textbf{13.35} & \textbf{18.13} & \textbf{22.25} & \textbf{29.26} \\
    \bottomrule
\end{tabular}
\end{center}
\vspace{-0.4cm}
\caption{Quantitative evaluation of \textbf{targeted} attack on CIFAR-10. We report the ASR for varying constraints on sparsity `$k$' and $\ell_{\infty}$-norm of the magnitude of perturbation `$\epsilon$'.}
\label{tab:results-cifar-tar}
\end{table*}

\end{document}